\documentclass[letterpaper]{article} 
\usepackage{aaai2026}  
\usepackage{times}  
\usepackage{helvet}  
\usepackage{courier}  
\usepackage[hyphens]{url}  
\usepackage{graphicx} 
\usepackage{natbib}  
\usepackage{caption} 
\usepackage{algorithm}
\usepackage{algorithmic}
\usepackage{amsmath}
\usepackage{booktabs}
\usepackage{newfloat}
\usepackage{listings}
\DeclareCaptionStyle{ruled}{labelfont=normalfont,labelsep=colon,strut=off} 
\floatstyle{ruled}
\newfloat{listing}{tb}{lst}{}
\floatname{listing}{Listing}
\title{AI Training Manager: Bounded Closed-Loop Control of Adaptive Training Recipes}
\author {
    Anjali Rao\textsuperscript{\rm 1},
    Nikhil Kamalkumar Advani\textsuperscript{\rm 1}
}
\affiliations {
    \textsuperscript{\rm 1}Independent Researcher\\
    10.anjalirao@gmail.com, nik.advani16@gmail.com
}

\usepackage{bibentry}

\begin{document}

\maketitle

\begin{abstract}
We present the AI Training Manager, a bounded LLM-based metacognitive monitoring-and-control layer for machine-learning training. The Manager asynchronously observes structured telemetry from an active training run, assesses the current training regime, and selects adaptive interventions through a constrained, deterministically verified action interface. We evaluate the feasibility of this architecture using deliberately induced, interpretable failure modes in supervised and reinforcement learning tasks. For supervised learning, we induce a multi-objective overfitting failure in a GPT-2-style model trained on TinyStories. The manager prevents the resulting late-run validation collapse, reducing the final validation loss by 54.3\% relative to the same stressed recipe. For reinforcement learning, we study robotic reaching under opposing multifactor stress regimes. In a conservative regime, the manager raises final deterministic safe success from 0.413 to 0.705. In an aggressive regime, it raises the safe success from 0.121 to 0.764. The same manager instructions and action surface are used in both regimes. These results suggest that bounded LLM reasoning can provide a practical metacognitive control layer over ongoing learning processes.
\end{abstract}


\section{Introduction}

Training a machine-learning model requires choosing a training recipe: optimizer
settings, learning-rate schedules, regularization, loss or reward weights,
exploration settings, and other controls that shape how learning proceeds.
Some of these controls remain fixed, while others may follow predefined
schedules or feedback rules. In either case, choices that are reasonable when
a run begins may become poorly matched to its evolving state. A model may
begin to overfit, one objective may saturate while another deteriorates, or an RL agent may explore too little to learn effectively, or so aggressively that unsafe behavior impedes learning. Such regimes can involve several interacting
training controls and may require different, sometimes opposing, corrective
responses. 

Discovering such problems late can be expensive. Large perception, language,
and embodied-learning workloads may require substantial compute time, so
an unsuitable training recipe can waste days of computation before
its effects become clear. The usual response is then to diagnose the run,
modify its training controls, and restart or resume from a checkpoint. This
motivates a supervisory mechanism that can instead inspect an active run,
reason about its evolving training state, and make bounded corrective
interventions while training is still underway.

\begin{figure}[t]
\centering
\includegraphics[width=\columnwidth]{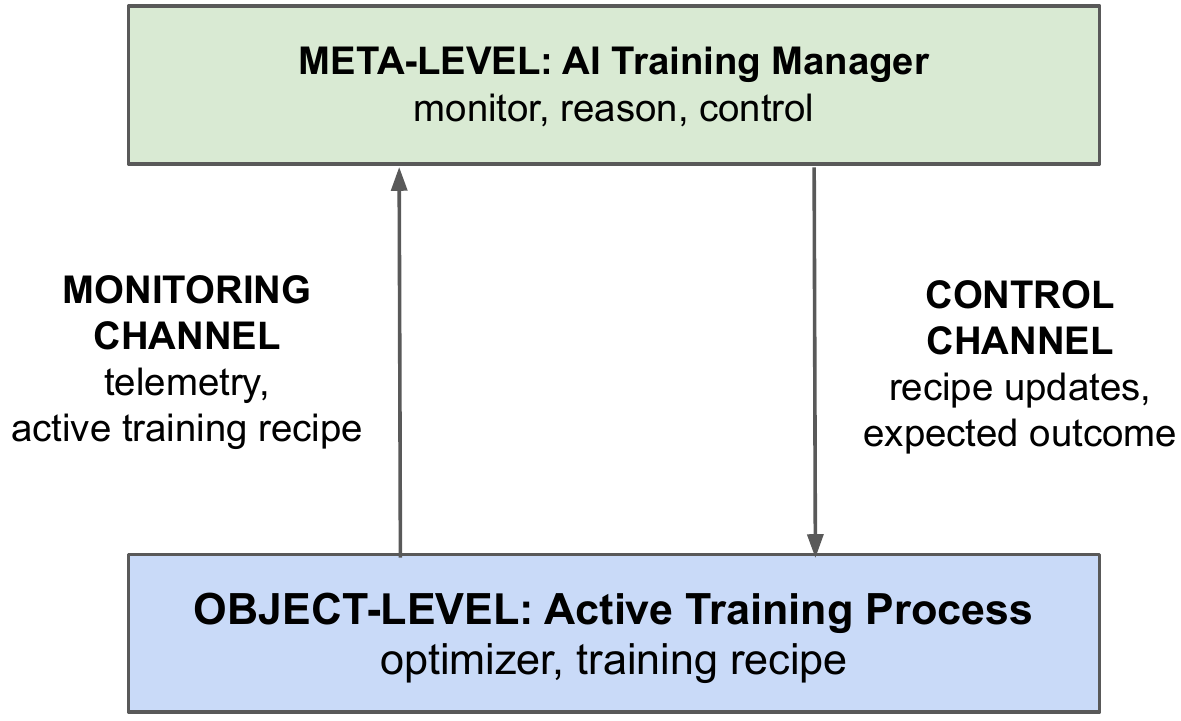}
\caption{\textbf{Training as a metacognitive monitoring-and-control problem.}
The active training process forms the object level, the AI Training Manager
operates at the meta level, telemetry provides the monitoring channel, and 
bounded updates provide the control channel.}
\label{fig:metacog-overview}
\end{figure}

We formulate this form of in-run intervention as a metacognitive
monitoring-and-control problem. In the classical two-level account of
metacognition, an object-level process is observed through a monitoring
channel, while control from a meta level regulates that process
\cite{nelson1990metamemory}. Here, the optimizer and environment form
the \emph{object-level}; structured telemetry provides \emph{monitoring}; bounded changes
to the active training recipe provide \emph{control}. The \emph{meta-level} is executed by the AI Training Manager: an LLM-based supervisor that monitors an
ongoing run, assesses its current training regime, selects a bounded
intervention, and subsequently observes the consequences of that intervention.
We use \emph{metacognitive} in this functional, system-level sense and do not
claim introspective awareness by the underlying learned model.

At periodic monitoring points, the Manager receives structured
telemetry, the current training recipe, recent evaluations, and intervention
history. It reasons over this evidence and selects an intervention from a
predefined action surface. Crucially, the Manager does not modify learned model
parameters directly; model weights continue to be updated by the ordinary
optimizer. Instead, it modifies only exposed training controls, such as
learning rates, loss or reward weights, regularization, exploration settings,
and optimizer configuration. Proposed actions must satisfy deterministic
schema and bounds checks before execution. Manager inference proceeds
asynchronously with object-level optimization, allowing reasoning and
intervention to occur during the active run rather than only between
experiments. The Manager also records the consequence it expects from each
intervention, so subsequent observations can inform later meta-level
decisions. The central idea is to replace a collection of failure-specific
state-to-action rules with a common supervisory loop that can interpret
heterogeneous evidence and choose among bounded controls according to the
current training regime.

We evaluate the approach on supervised and reinforcement
learning with three independent seeds per condition. On a TinyStories ~\cite{eldan2023tinystories}
overfitting stress test, the AI Training Manager reduces final validation loss by 54.3\%. 
We then evaluate randomized robotic
reaching under opposing multifactor stress regimes spanning exploration,
reward shaping, and policy-update dynamics. The same manager instructions,
action surface, model configuration, and decision procedure are used in both
RL regimes. The Manager increases final deterministic safe success from
\(0.413\) to \(0.705\) in the conservative regime and from
\(0.121\) to \(0.764\) in the aggressive regime.

Our contributions are:
\begin{itemize}
    \item A metacognitive formulation of adaptive training in which an ongoing
    machine learning training process constitutes the object level of an explicit
    monitoring--reasoning--control loop, with prior interventions and their
    observed consequences fed back into subsequent meta-level assessments.

    \item A bounded-authority 'AI Training Manager' architecture in which an LLM
    reasons over heterogeneous training telemetry and intervention history,
    while a predefined action surface and deterministic verification constrain
    executable control; meta-level reasoning runs asynchronously with
    object-level learning.

    \item An empirical evaluation across supervised and reinforcement learning
    using controlled failure injection and three seeds per condition,
    including opposing multi-factor RL regimes, to test whether the same
    meta-level supervisor can identify and recover qualitatively different
    training states.
\end{itemize}

\section{Related Work}

\subsection{Artificial Metacognition and Runtime Monitoring}

Metacognition is commonly characterized by a distinction between an
object-level process and a meta-level, connected through monitoring and control:
information about the object level informs meta-level assessment, while
control regulates the underlying process \cite{nelson1990metamemory}.
Computational metacognition extends this organization to artificial systems
\cite{cox2022computational}, and recent work similarly emphasizes monitoring an
AI system's operation and using those observations to regulate subsequent
behavior \cite{shakarian2026toward,laird2025cmcmetacognition}. Classical
architectures such as MIDCA also realize the meta level as a distinct component
that monitors and regulates object-level reasoning.

Runtime failure monitoring provides a complementary perspective on observing the behavior of AI systems. Prior work uses internal or task-relevant signals to detect failures in perception, trajectory prediction, and online map or trajectory generation \cite{antonante2023monitoring,farid2023task,advani2026auditing}. While these approaches primarily monitor deployed inference, we treat the ongoing learning process itself as the object level and couple monitoring with bounded corrective control over training.

\subsection{Adaptive Training and Dynamic Configuration}

Machine-learning training already includes many forms of adaptation, including
learning-rate schedules, early stopping, curriculum learning, and adaptive
loss weighting
\cite{smith2017cyclical,prechelt1998early,bengio2009curriculum,
kendall2018multi,chen2018gradnorm}. Hyperparameter-optimization methods such as
Bayesian optimization, Hyperband, and Optuna typically select configurations
across trials
\cite{snoek2012practical,li2018hyperband,akiba2019optuna}, while
Population-Based Training adapts hyperparameters during training across a
population of learners \cite{jaderberg2017population}.

Dynamic algorithm configuration (DAC) formulates
runtime parameter selection as a sequential control problem in which a policy
observes an algorithm's state and changes its configuration during execution
\cite{adriaensen2022automated}. The AI Training Manager fits this broad
formulation, but uses a pretrained LLM at decision time rather than a dedicated
learned state-to-parameter policy. It reasons over heterogeneous telemetry,
the current training recipe, and intervention history, while a predefined
action surface and deterministic verifier constrain executable changes. Our
contribution relative to DAC is therefore not online adaptation itself, but an
LLM-based, inspectable, bounded-authority realization of dynamic training
control.

\begin{figure*}[!t]
    \centering
    \includegraphics[width=0.92\textwidth]{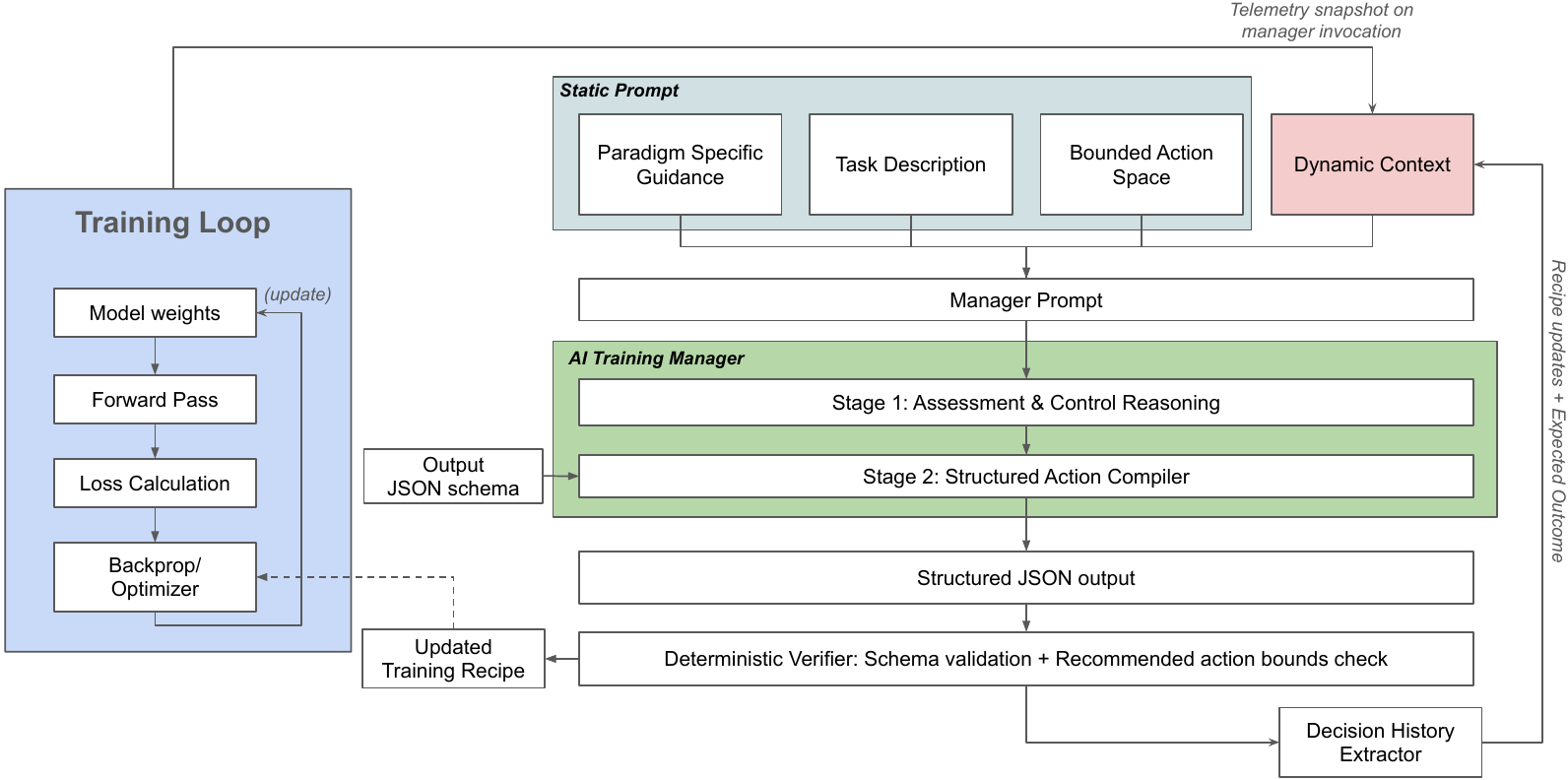}
    \caption{\textbf{AI Training Manager architecture.}
    The active training process is monitored through structured telemetry and
    dynamic run context. A two-stage manager separates meta-level assessment
    and control reasoning from structured action compilation; a deterministic
    verifier restricts execution to the predefined action surface. Verified
    recipe updates are applied asynchronously at admissible training
    boundaries, and their observed outcomes are returned through decision
    history.}
    \label{fig:training-manager}
\end{figure*}

\subsection{LLM-Based Optimization and Automated Experimentation}

LLMs have increasingly been used for optimization and experimentation. OPRO
proposes candidate solutions from previous solutions and scores
\cite{yang2023large}; AgentHPO configures and interprets successive
hyperparameter trials \cite{liu2025agenthpo}; and systems such as
MLAgentBench and The AI Scientist automate broader experimental and research
workflows \cite{huang2024mlagentbench,lu2024aiscientist}. These approaches
primarily operate across trials or experiments. The Training Manager instead
regulates an experiment that is already running, without selecting a new
experiment or rewriting the training program.

AutoEP is the closest LLM-based analogue. It uses quantitative descriptions of
a running metaheuristic together with LLM reasoning to adapt hyperparameters
online \cite{xu2026autoep}. We similarly use a pretrained LLM for
state-dependent online control, but the controlled process and evaluation
setting differ. The Training Manager regulates machine-learning training using
signals about optimization, generalization, task performance, and safety, and
can modify multiple training controls through a verified action interface. Our
evaluation focuses on whether the same meta-level controller can recognize and
recover qualitatively different training regimes, including opposing
multifactor failures.

\section{AI Training Manager}
\label{sec:method}

The AI Training Manager augments a conventional machine-learning training
process with a distinct meta-level controller. As illustrated in
Figure ~\ref{fig:metacog-overview}, the active training process forms the object
level, telemetry provides a monitoring channel to the Manager, and bounded
interventions by the Manager provide a control channel back to training. The object-level
optimizer and learning algorithm continue to perform the actual parameter
updates. The Manager reasons about the evolving training process and
can modify only a predefined set of training controls. Subsequent outcomes are
returned to the manager at later monitoring points, forming a recurrent
monitoring--reasoning--control loop.

\subsection{Monitoring Context}

At each monitoring point, the Manager receives a structured context describing
both the training problem and the current state of the run. This context
combines three static components (paradigm-specific guidance, a task
description, and the intervention surface available to the manager) with
dynamic context from the active training process.

The paradigm-specific guidance provides general principles for interpreting
training behavior within the learning paradigm; the task description specifies
the concrete objective, success and failure criteria; and the intervention surface specifies which controls are available
to the manager and the forms of modification permitted. These static
components are fixed for a given experimental setup. 

The dynamic context includes the current training recipe, current performance
and optimization telemetry, recent evaluation history, checkpoint state, run
progress, and a compact history of previous manager decisions and their
subsequently observed outcomes. Before prompt construction, trainer state is
canonicalized into a single representation so that duplicated or invalid
measurements do not receive disproportionate weight. Decision history records the prior assessment, intended intervention,
expected outcome, and measurements subsequently observed. This feedback makes the meta-level process recurrent across decision
points rather than a sequence of independent LLM calls.

\subsection{Meta-Level Reasoning and Control}

\paragraph{Two-stage decision pipeline:}
Figure~\ref{fig:training-manager} shows the implementation of the meta-level
controller. We separate substantive reasoning about the training process from
the mechanical generation of an executable action. Let $g$ denote
paradigm-specific guidance, $q$ the task description, $c_t$ the dynamic
monitoring context at decision point $t$, $\mathcal{A}_t$ the currently
available intervention surface, and $\mathcal{S}$ the executable output
schema. The pipeline is
\begin{align}
z_t &= M_{\mathrm{reason}}(g,q,\mathcal{A}_t,c_t), \\
\hat a_t &= M_{\mathrm{compile}}(z_t,\mathcal{A}_t,\mathcal{S}), \\
a_t &= V(\hat a_t,\mathcal{A}_t,\mathcal{S}),
\end{align}
where $z_t$ contains the Stage~1 assessment and intended control decision,
$\hat a_t$ is its structured representation, and $V$ is the deterministic
verifier. Stage~1 receives the full monitoring context and determines whether
an intervention is warranted. When acting, it selects an intervention and
records its expected consequence and the measurements that should be examined
subsequently. A no-action decision is always available. Stage~2 functions as an action compiler rather than a second diagnostic
agent. It receives the completed Stage~1 decision, the currently available
intervention surface, and the required output schema, but no independent
training telemetry or history. It is instructed to preserve the substantive
decision made by Stage~1 rather than re-diagnose the run or choose a different
intervention. 

This decomposition separates semantic reasoning from reliable
machine-readable generation. It is a pragmatic implementation choice rather
than a requirement of the metacognitive formulation. A sufficiently reliable
model could in principle perform both functions in a single pass.

\paragraph{Model configuration:}
We use GPT-5.4 at temperature $0.6$ for Stage~1, whose role requires reasoning
over heterogeneous evidence and selecting among alternative interpretations
and interventions. Stage~2 uses GPT-5.4-mini at temperature $0.1$, reflecting
its narrower role as a structured action compiler. 

\paragraph{Bounded intervention surface:}
We define an intervention surface before the training run that
specifies which elements of the training process may be modified and the
authority granted over each one. The exposed controls are existing elements
of the training recipe, such as optimizer or objective parameters appropriate
to that paradigm. Learned model weights, model architecture, training code,
and the task objective itself remain outside the manager's authority. Within
a paradigm, the same intervention surface is used across the experimental
conditions being compared.

The numerical authority granted over an exposed control is bounded at two
distinct levels. First, each control has an absolute admissible range that
limits the values it may take over the course of training. Second, each
manager intervention is subject to a step bound limiting how far that control
may move from its current value in a single decision. Thus, a value can lie
within the global admissible range while still being unreachable through one
intervention. This enforces gradual adaptation rather than permitting an
arbitrarily large one-step recipe change. The currently available surface
$\mathcal{A}_t$ may be narrower still because some interventions have
state-dependent eligibility conditions.

This separation between reasoning and executable authority reflects a
deliberate separation of concerns. The
LLM determines what it believes should change, while the surrounding system
determines what it is capable of changing and by how much. Bounding authority
limits the consequences of an incorrect meta-level judgment, preserves the
distinction between supervision and object-level learning, and makes
interventions more reproducible and auditable. These constraints are not
intended as a formal safety guarantee.

\paragraph{Deterministic verification:}
The deterministic verifier checks that the compiled intervention conforms to
the required schema, refers only to permitted controls or operations, remains
within the corresponding absolute parameter bounds, respects the permitted
per-intervention step limits, and satisfies any applicable state-dependent
eligibility constraints. Invalid outputs receive one formatting retry and
otherwise resolve to no action. The verifier therefore enforces the
boundary of the manager's executable authority; it does not determine whether
the Stage~1 assessment is correct, whether the intervention is optimal, or
whether an otherwise admissible action will improve training.

\paragraph{Rollback and state-dependent recovery:}
In addition to recipe modifications, the intervention surface may contain
discrete state-recovery operations. In the reinforcement-learning experiments,
the manager can roll back to a previously saved policy and optimizer
checkpoint after sustained performance regression. Rollback restores learned
and optimizer state while retaining the manager's current training recipe, so
the recovery operation does not reinstate the configuration that may have
produced the regression. Checkpoint selection, rollback eligibility, and
cooldown rules are specified in the reinforcement-learning setup.

\paragraph{Asynchronous application and feedback:}
Manager requests are initiated ahead of scheduled intervention boundaries, allowing object-level optimization to continue while the manager reasons. Each request is associated with a bounded response window. A response received within this window is verified and applied at an admissible training boundary rather than interrupting an ongoing optimization step or episode. Before application, the trainer obtains its current state and rechecks state-dependent eligibility, accounting for the fact that the manager's decision was formed using an earlier telemetry snapshot. 

After an intervention is applied, subsequent measurements and the resulting
recipe are associated with the corresponding manager decision and returned
through the decision-history channel at later monitoring points. The resulting
monitoring $\rightarrow$ reasoning $\rightarrow$ bounded control
$\rightarrow$ observation cycle closes the metacognitive feedback loop used
throughout both experiments.

\begin{figure*}[t]
    \centering
    \includegraphics[width=\textwidth]{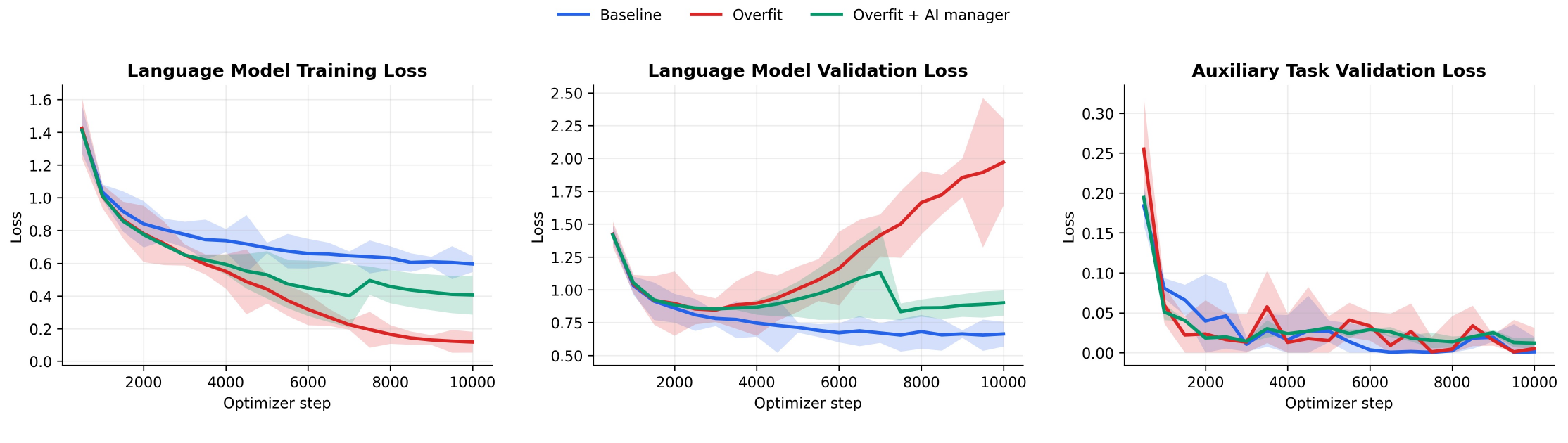}
    \caption{\textbf{TinyStories overfitting experiment.}
Language-model training loss (left), validation loss (center), and auxiliary dialogue-classification validation loss (right) over $10{,}000$ optimizer steps. Curves show the mean across three independent seeds; shaded regions indicate ±1 standard deviation. While the fixed overfit recipe continues reducing training loss at the expense of validation performance, manager-controlled training substantially mitigates this generalization failure.}
    
    \label{fig:tiny-results}
\end{figure*}

\section{Experiments and Results}
\label{sec:experiments}

We evaluate the AI Training Manager in two learning settings using deliberately induced, interpretable failure modes. These controlled experiments allow the manager's diagnosis, intervention selection, and subsequent training response to be directly inspected. Our objective is to demonstrate the feasibility of a common supervisory architecture across distinct training regimes, rather than establish comparative superiority over specialized adaptive training methods designed for individual failure modes. 

We first study supervised autoregressive language-model training on
TinyStories, using a 
multi-objective overfitting stress test. We then study reinforcement learning
for randomized robotic reaching under opposing multifactor stress regimes.
Each experimental condition is run with
3 random seeds and results are reported as mean $\pm$ standard
deviation.

\subsection{TinyStories: Multi-Objective Overfitting}
\label{sec:supervised}

\paragraph{Experimental setup:}
We train a small
GPT-2-style autoregressive Transformer with six Transformer layers, six
attention heads, embedding dimension $384$, and context length $256$.
Training runs for $10{,}000$ optimization steps using AdamW, batch size $16$
with two gradient-accumulation steps, gradient clipping at $1.0$, and cosine
learning-rate annealing from $10^{-3}$ to $10^{-4}$. 
Evaluation is performed every $500$ optimization steps. 

Manager
requests are dispatched $100$ steps before each scheduled
evaluation (e.g., step numbers $400$, $900$, etc.). A response
is considered temporally valid if it arrives within $500$
steps of its corresponding request. Valid responses are
verified and applied at the next admissible optimization-step
boundary, while responses arriving after this window are
discarded as stale. Training continues asynchronously during
manager inference.

In addition to next-token prediction, we attach an auxiliary binary
classification head to the model whose task is to predict whether a story contains dialogue. The model is trained jointly on both the objectives:
\begin{equation}
    \mathcal{L}
    =
    \mathcal{L}_{\mathrm{LM}}
    +
    \lambda_{\mathrm{aux}}\mathcal{L}_{\mathrm{aux}},
\end{equation}
where $\mathcal{L}_{\mathrm{LM}}$ is the autoregressive language-modeling
loss, $\mathcal{L}_{\mathrm{aux}}$ is the dialogue-classification loss, and
$\lambda_{\mathrm{aux}}$ controls the contribution of the auxiliary task.

We compare three conditions. The \emph{Baseline} trains on $20{,}000$
stories using the fixed cosine learning-rate schedule, and $\lambda_{\mathrm{aux}}=0.1$. The \emph{Overfit}
condition deliberately makes generalization more difficult by reducing the
training set to $2{,}000$ stories while retaining the same initially
unregularized recipe and auxiliary objective. Its recipe remains fixed
throughout training. The \emph{Overfit + Manager} condition begins from the
identical overfit configuration, but exposes learning rate, dropout, weight
decay, and $\lambda_{\mathrm{aux}}$ to the Training Manager.

\paragraph{Generalization recovery:}
The manager strengthens regularization, increasing dropout to 0.20 and weight decay to 0.0293, while adjusting the learning-rate trajectory to values as low as $7.43\times10^{-5}$ (Table~\ref{tab:tiny-controls}).

Figure~\ref{fig:tiny-results} shows a clear separation between optimization
of the training objective and generalization in the \emph{Overfit} condition.
Its final language-model training loss falls to $0.118\pm0.064$,
substantially below both the \emph{Baseline} and \emph{Overfit + Manager} . This does not translate into
better validation performance.  Instead, \emph{Overfit} validation loss initially
improves and then rises sharply, reaching $1.971\pm0.331$ at the end of
training. \emph{Overfit + Manager}  has a final validation
loss of $0.900\pm0.096$, a $54.3\%$ reduction relative to \emph{Overfit}. The
\emph{Baseline} finishes at $0.663\pm0.095$. Thus, manager control does not fully eliminate the generalization disadvantage associated with the much smaller $2{,}000$-story training set, but it prevents most of the late-run validation collapse. A conventional early-stopping rule could also have mitigated this deliberately induced failure by selecting an earlier checkpoint. The purpose of this experiment, however, is not to demonstrate superiority over early stopping, but to provide an interpretable test of whether a general supervisory architecture can recognize deteriorating generalization and adapt multiple training controls while learning continues.

\paragraph{Multi-objective regulation:}
The auxiliary dialogue-classification task provides a distinct signal from the primary language-modeling objective. As shown in Fig.~\ref{fig:tiny-results}, auxiliary validation loss becomes small early in training and remains low even as language-model validation performance deteriorates.

The manager responds by reducing $\lambda_{\mathrm{aux}}$, decreasing the auxiliary objective's contribution to the joint training loss (Table~\ref{tab:tiny-controls}). This demonstrates objective-specific regulation: the manager adjusts the relative emphasis of an already well-performing secondary task while continuing to address generalization failure in the primary objective.

\begin{table}[t]
\centering
\small
\setlength{\tabcolsep}{4pt}
\caption{\textbf{TinyStories manager-controlled parameters.}
Baseline and Overfit use the same training recipe, differing only in
training-set size (20k versus 2k stories). Overfit + Manager starts
from the same configuration as Overfit and adaptively modifies the
exposed controls. Arrows indicate changes from the initial value to
the observed extremum across the three manager-controlled runs.}
\label{tab:tiny-controls}
\begin{tabular}{@{}lcc@{}}
\toprule
\textbf{Parameter} &
\shortstack{\textbf{Baseline / Overfit}} &
\shortstack{\textbf{Overfit + Manager}} \\
\midrule
Learning rate
    & $10^{-3} \rightarrow 10^{-4}$
    & $10^{-3} \rightarrow 7.43 \times 10^{-5}$ \\
Weight decay
    & $0$
    & $0 \rightarrow 0.0293$ \\
Dropout
    & $0$
    & $0 \rightarrow 0.20$ \\
$\lambda_{\mathrm{aux}}$
    & $0.1$
    & $0.1 \rightarrow 0.00419$ \\
\bottomrule
\end{tabular}
\end{table}


\paragraph{Manager decisions and Prospective Outcome Agreement:}
Before each intervention, the manager records an expected
consequence and evaluation horizon. A deterministic audit
script subsequently compares observed telemetry against
the pre-intervention reference over the specified horizon.
Each expectation is assessed along two dimensions:
whether the predicted control mechanism occurred and
whether the subsequent task-performance trajectory
was supported. An outcome is classified as \emph{fully observed} when
both criteria are satisfied, \emph{partially observed}
when exactly one is satisfied, and \emph{not observed}
when neither is satisfied. 

Across the three \emph{Overfit + Manager} runs, the manager was
issued 57 requests, resulting in 9 applied interventions
and 46 no-action decisions. 2 had an invalid-response resulting in a no-action fallback. Of 9 assessable prospective
expectations, 5 were fully observed, 3 were
partially observed, and 1 was not observed. None of the manager responses were stale.


Taken together, the TinyStories experiment provides an interpretable test of
the complete monitoring--reasoning--control loop. The manager observes a
primary objective developing a generalization failure while a secondary
objective is already largely solved and responds by modifying several
elements of the exposed training recipe. 
The experiment does not establish
superiority to a handcrafted controller designed specifically for this
failure; rather, it demonstrates that the common manager architecture can
infer the evolving regime from telemetry and substantially limit the resulting
generalization collapse. 




\subsection{Reinforcement Learning: Robotic Reaching}
\label{sec:rl}

\paragraph{Environment and Training Setup:}
We next evaluate the AI Training Manager in a continuous-control reinforcement
learning problem where qualitatively different training failures require
different, and in some cases opposing, interventions. The task is planar
robotic reaching with a two-link articulated arm. At the beginning of every
episode, the initial joint configuration, target position, circular-obstacle
position, and obstacle radius are randomized subject to feasibility
constraints. The policy observes a 14-dimensional state containing sine and cosine of both
joint angles, target and obstacle geometry, obstacle radius, end-effector
position, the vector from the end effector to the target, and current obstacle
clearance. Actions are two continuous joint-velocity commands  An episode terminates when the target is reached safely, a collision
occurs, or $120$ simulation steps have elapsed. Success is defined as
reaching the target without collision. The reward combines distance progress, a safe-success bonus, a collision
penalty, and fixed movement and clearance penalties. Importantly, changes to
reward coefficients alter only the training signal; the definitions of
success and collision used for evaluation remain unchanged.

We train using PPO with two-layer actor and critic MLPs with $128$
hidden units per layer and $\tanh$ activations. The Gaussian actor maintains a
learned log standard deviation. Training proceeds for $10^6$ environment interactions. Each seed run used an independently initialized manager with separate telemetry, checkpoints, and intervention history; no manager decisions or histories were shared across runs. We evaluate the learned policy using
deterministic safe success, collision rate, and final distance to goal.
During training, the manager received diagnostic evaluations comprising 80 episodes per difficulty (easy, medium, and hard) in both deterministic and stochastic modes. Final held-out evaluation used disjoint random seeds and 240 episodes per difficulty. These final evaluations were not available to the manager or used for training-time decisions or checkpoint selection.

Manager requests are dispatched $5$ PPO updates before each
scheduled intervention boundary (e.g., update $15$ for an
intended boundary at update $20$). A response is considered
temporally valid if it arrives within $10$ PPO updates of
its corresponding request. Valid responses are verified
against the current trainer state and applied at the next
admissible update boundary. Responses arriving after this
window are discarded as stale. PPO training continues
asynchronously without waiting for manager responses.

\paragraph{Code and reproducibility:}
The AI Training Manager implementation, manager prompts, action
verification, and reproduction
instructions are available at
\url{https://github.com/NikAdvani16/AI-Training-Manager}.

\paragraph{Controlled stress regimes:}
We construct two deliberately opposing multi-factor stress conditions around a
healthy PPO reference recipe. The stress conditions jointly perturb
exploration, reward shaping, and policy-update dynamics rather than changing a
single hyperparameter in isolation. The \emph{aggressive} regime jointly increases policy
update magnitude and exploration pressure while weakening collision aversion
and strengthening progress incentives. The \emph{conservative} regime
instead restricts policy updates, suppresses exploration, increases collision
penalties, and weakens goal-directed incentives. These configurations are
intended as stress tests of meta-level regulation rather than causal
ablations of individual hyperparameters. The same manager instructions,
model configuration, action surface, and decision procedure are used in both
regimes; only the observed telemetry and intervention history differ. The parameters chosen for these can be found in Table~\ref{tab:rl-controls}.  

\begin{figure*}[t]
    \centering
    \includegraphics[width=\textwidth]{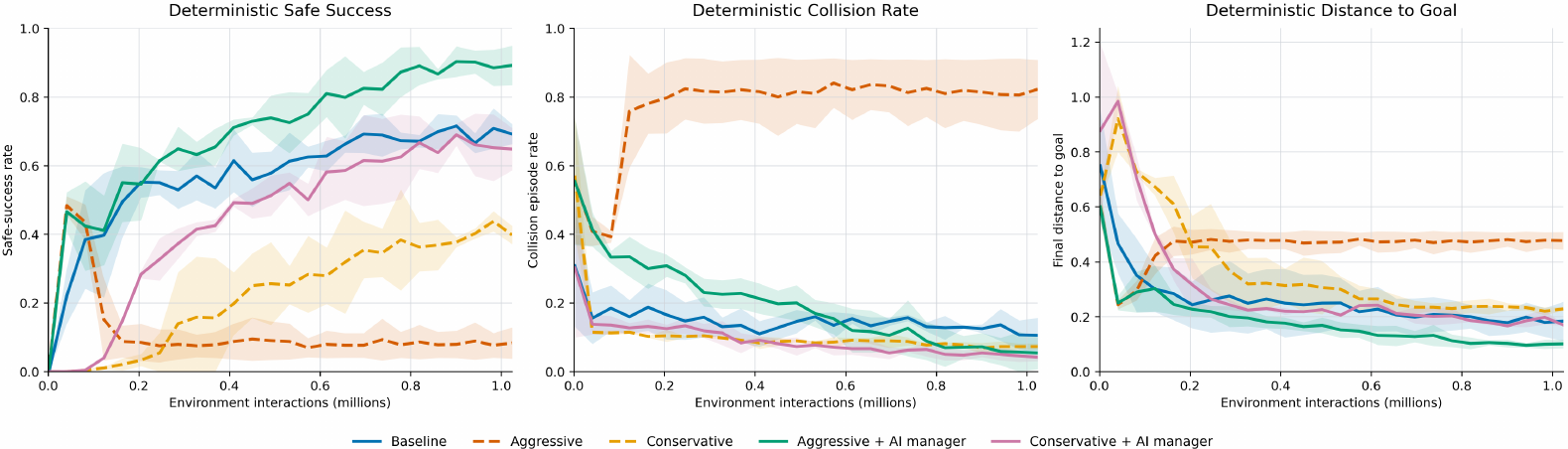}
    \caption{\textbf{Recovery from opposing reinforcement-learning stress regimes.}
    Deterministic safe success (left), collision rate (center), and final
    distance to goal (right) over $10^6$ environment interactions.
    Curves show the mean across three independent seeds and shaded regions
    denote $\pm1$ standard deviation. The fixed aggressive recipe becomes
    collision dominated, whereas the fixed
    conservative recipe learns
    the reaching task slowly and incompletely. The Manager recovers both
    regimes: it suppresses unsafe behavior in the aggressive condition
    while restoring task progress in the conservative condition. Curves represent training-time diagnostic evaluations; final held-out performance is reported separately.}
    \label{fig:rl-results}
\end{figure*}

\paragraph{Recovery under opposing failures:}
Figure~\ref{fig:rl-results} shows that the two stress recipes produce
qualitatively different failure signatures. The fixed aggressive policy
initially makes progress but rapidly enters a collision-dominated regime:
collision rate rises above $0.8$ and remains high, deterministic safe success
stays near $0.1$, and final distance to goal plateaus well above the healthy
baseline. In contrast, the conservative condition maintains a low
collision rate but learns much more slowly; its safe success remains
substantially below the healthy baseline and its distance-to-goal trajectory
shows that avoiding collisions alone does not constitute successful task
learning.

The Manager recovers useful learning in both cases despite receiving
the same static instructions and control authority. In the aggressive
condition, manager control reverses the collision-dominated trajectory.
Collision rate falls throughout training while safe success and goal-reaching
performance improve. In the
conservative condition, manager control restores task progress while
retaining the low-collision behavior of the original regime. On the final
held-out evaluation, deterministic safe success increases from
$0.121\pm0.019$ to $0.764\pm0.037$ in the aggressive regime and from
$0.413\pm0.036$ to $0.705\pm0.021$ in the conservative regime, compared
with $0.695\pm0.037$ for the healthy baseline. Over the complete training
trajectory, the manager reduces collision-rate AUC by $67.5\%$ in the
aggressive regime and more than doubles deterministic safe-success AUC
in the conservative regime.

\paragraph{Different interventions for opposing failure modes:}
The recovery is not produced by simply restoring both runs to the healthy
recipe. Table~\ref{tab:rl-controls} summarizes the parameter values reached
under manager control. In the aggressive regime, the manager sharply
reduces actor learning rate and entropy pressure, increases the collision
penalty, and partially tightens the KL constraint, while retaining several
other aggressive settings. In the conservative regime, it instead
increases policy-update capacity and the safe-success incentive while
preserving strong collision aversion. Notably, entropy is not simply increased
in the conservative condition. Thus, the two regimes elicit distinct
multi-axis control strategies rather than symmetric movement toward a
predefined baseline recipe.

\begin{table*}[t]
\centering
\small
\caption{\textbf{Training recipe control in reinforcement learning.}
The baseline recipe and initial stress configurations are fixed. Ranges denote
the values observed after AI Training Manager interventions across the
manager-controlled runs. }
\label{tab:rl-controls}
\begin{tabular}{lccccc}
\toprule
\textbf{Parameter} &
\textbf{Baseline} &
\textbf{Aggressive} &
\textbf{Aggressive + Manager} &
\textbf{Conservative} &
\textbf{Conservative + Manager} \\
\midrule
Actor learning rate
  & $3{\times}10^{-4}$ & $3{\times}10^{-3}$
  & $1.88{\times}10^{-4}$--$9.49{\times}10^{-4}$
  & $3{\times}10^{-5}$
  & $2.15{\times}10^{-4}$--$9.35{\times}10^{-4}$ \\
Critic learning rate
  & $1{\times}10^{-3}$ & $3{\times}10^{-3}$
  & $3{\times}10^{-3}$
  & $1{\times}10^{-3}$
  & $1{\times}10^{-3}$--$1.5{\times}10^{-3}$ \\
Entropy coefficient
  & $0.01$ & $0.08$
  & $0.00141$--$0.005$
  & $0.001$
  & $0.00025$--$0.001$ \\
PPO clip coefficient
  & $0.20$ & $0.30$
  & $0.25$--$0.30$
  & $0.10$
  & $0.10$--$0.12$ \\
PPO update epochs
  & $6$ & $8$
  & $8$
  & $3$
  & $3$ \\
Target KL
  & $0.03$ & $0.10$
  & $0.05$--$0.10$
  & $0.01$
  & $0.02$--$0.05$ \\
Collision penalty
  & $5$ & $1$
  & $10$
  & $25$
  & $25$--$35$ \\
Safe-success bonus
  & $5$ & $10$
  & $10$
  & $2.5$
  & $3$--$5$ \\
Progress-reward scale
  & $3$ & $5$
  & $5$
  & $1.5$
  & $1.5$ \\
\bottomrule
\end{tabular}
\end{table*}

\paragraph{Manager decisions and Prospective Outcome Agreement:}

Using the deterministic audit procedure described in the
TinyStories experiment, we assess whether the manager's
prospective expectations are supported by subsequent telemetry.
Across the six manager-controlled RL runs, 144 manager requests produced 86 applied interventions,
 and 58 no-action decisions. No requests produced a terminal invalid-response fallback. Of 86 assessable prospective expectations, 64 ($74.4\%$) were fully observed, 21 ($24.4\%$) were partially observed, and 1 ($1.2\%$) was not observed. None of the manager responses were stale. Because a single manager
decision may change multiple controls whose expected effects are evaluated
against the same subsequent trajectory, these observations are not
statistically independent; we report them as a descriptive measure of
prospective outcome agreement rather than prediction accuracy or calibration. Checkpoint rollback was also available through the same bounded action
surface. It was invoked
only once.

\paragraph{Intervention-history ablation.}
To probe the role of intervention history, we ran an additional
aggressive-regime experiment in which the Manager received the
current telemetry and training recipe but not previous intervention
decisions, expected outcomes, or their observed consequences. This
history-ablated variant retained the same instructions and
action surface.

Across 24 decisions, the history-ablated Manager intervened at every
opportunity. By comparison, across the three standard Manager runs in
the aggressive regime, the Manager averaged 14 interventions and
10 \texttt{no\_action} decisions per run. Without intervention history,
the Manager repeatedly reduced entropy and increased the collision
penalty before the effects of earlier changes could be evaluated.
The single history-ablated run achieved a final held-out deterministic
safe success of 0.656, compared with $0.764 \pm 0.037$ for the Manager
across three seeds.

The ablation complements the opposing-regime experiments by probing a
different aspect of the closed loop. The opposing stress regimes show
that current telemetry can elicit qualitatively different control
strategies under identical Manager instructions, while the history
ablation suggests that feedback about previous interventions helps
moderate subsequent control rather than repeatedly acting on the
current state alone. Taken together, these results provide evidence
for both state-dependent monitoring and feedback-informed control
within the proposed metacognitive loop.

\section{Discussion and Limitations}
\label{sec:discussion}
Across supervised and reinforcement learning, the AI Training
Manager mitigated deliberately induced failures through
bounded, state-dependent interventions. Under opposing
RL stress regimes, identical manager instructions elicited
distinct corrective strategies, restoring final safe-success
performance to levels comparable to the healthy reference. 


\paragraph{When is a general reasoner useful?}
When a failure mode, its diagnostic statistic, and its corrective response are
known in advance, a handcrafted controller is simpler, cheaper, and likely
more reliable than an LLM. The motivation for a general meta-level reasoner
arises when several plausible failure hypotheses interact across heterogeneous
telemetry and overlapping controls, making it increasingly cumbersome to
specify a separate detector, repair policy, and arbitration rule for every
possible regime. We hypothesize that the potential advantage of general meta-level reasoning may therefore be
greatest in long-running, multi-objective training workloads where many
signals and controllable quantities evolve simultaneously. Examples include
mixture-of-experts language models with multiple losses, and large embodied systems such as end-to-end driving. 

\paragraph{Limitations:}
The present experiments are controlled, small-scale demonstrations rather than
large-scale deployments. We evaluate two learning paradigms with three seeds
per condition, and the failure modes are deliberately constructed stress
regimes chosen to make the resulting monitoring, intervention, and recovery
behavior interpretable. They therefore do not characterize the distribution
of naturally occurring failures in large training runs. We also do not compare against an exhaustive handcrafted controller or a separately trained DAC policy. The experiments therefore establish that the proposed architecture can perform state-dependent recovery, not that an LLM reasoner is always preferable to specialized control.

The manager's flexibility is also conditional on the action surface supplied
by the experimenter: it can select among and parameterize permitted
interventions, but cannot recover from a failure whose necessary corrective
mechanism is absent from that surface.

More importantly, the
experiments do not yet test the regime in which we hypothesize that a general
reasoner will be most valuable: training processes with substantially more
objectives, telemetry channels, interacting failure modes, and available
controls. Evaluating the architecture on such workloads is
the principal next step.

\section{Conclusion}
\label{sec:conclusion}

We presented the AI Training Manager, a bounded LLM-based metacognitive
monitoring-and-control layer for active machine-learning training. The manager
observes structured evidence from an ongoing run, forms an explicit
meta-level assessment, selects interventions through a constrained action
surface, and incorporates their observed consequences into subsequent
decisions. Across supervised learning on TinyStories and robotic reinforcement
learning, the same architecture recovered deliberately induced training
failures requiring qualitatively different control strategies. In
reinforcement learning, identical manager instructions produced distinct
multi-axis interventions under opposing multifactor stress regimes, while the
manager's prospective expected outcomes were generally reflected in subsequent
telemetry. These results provide initial evidence that bounded LLM reasoning
can serve as a practical meta-level supervisory mechanism when useful
intervention depends on interpreting the evolving state of the learning
process rather than applying a single predefined feedback rule.

\bibliography{aaai2026}

\end{document}